\UseRawInputEncoding
\pdfoutput=1
\documentclass[sigconf]{acmart}
\AtBeginDocument{%
  }

\usepackage{multirow}

\begin{document}

\title{LLM-EvRep: Learning an LLM-Compatible Event Representation Using a Self-Supervised Framework}

\author{Zongyou Yu}
\orcid{0009-0005-1098-9793}
\affiliation{%
  \institution{Beijing Technology and Business University}
  \city{Beijing}
  \country{China}}
\email{zongyou.yu@st.btbu.edu.cn}

\author{Qiang Qu}
\orcid{0000-0002-6648-5050}
\affiliation{%
  \institution{The University of Sydney}
  \city{Sydney}
  \country{Australia}}
\email{vincent.qu@sydney.edu.au}

\author{Qian Zhang}
\orcid{0009-0007-4824-3158}
\affiliation{%
  \institution{Beijing Technology and Business University}
  \city{Beijing}
  \country{China}}
\email{qian.zhang@st.btbu.edu.cn}

\author{Nan Zhang}
\orcid{0000-0003-4904-7857}
\affiliation{%
  \institution{Beijing Technology and Business University}
  \city{Beijing}
  \country{China}}
\email{20210504@btbu.edu.cn}

\author{Xiaoming Chen}
\authornote{Corresponding author.}
\orcid{0000-0002-7503-3021}
\affiliation{%
  \institution{Beijing Technology and Business University}
  \city{Beijing}
  \country{China}}
\email{xiaoming.chen@btbu.edu.cn}

\renewcommand{\shortauthors}{Z. Yu et al.}

\begin{abstract}
    Recent advancements in event-based recognition have demonstrated significant promise, yet most existing approaches rely on extensive training, limiting their adaptability for efficient processing of event-driven visual content. Meanwhile, large language models (LLMs) have exhibited remarkable zero-shot capabilities across diverse domains, but their application to event-based visual recognition remains largely unexplored. To bridge this gap, we propose \textbf{LLM-EvGen}, an event representation generator that produces LLM-compatible event representations \textbf{LLM-EvRep}, thereby enhancing the performance of LLMs on event recognition tasks. The generator is trained using a self-supervised framework, aligning the generated representations with semantic consistency and structural fidelity. Comprehensive experiments were conducted on three datasets: N-ImageNet, N-Caltech101, and N-MNIST. The results demonstrate that our method, \textbf{LLM-EvRep}, outperforms the event-to-video method, E2VID, by 15.93\%, 0.82\%, and 50.21\%, respectively, in recognition tasks when evaluated using GPT-4o.
\end{abstract}


\begin{teaserfigure}
  \includegraphics[width=\textwidth]{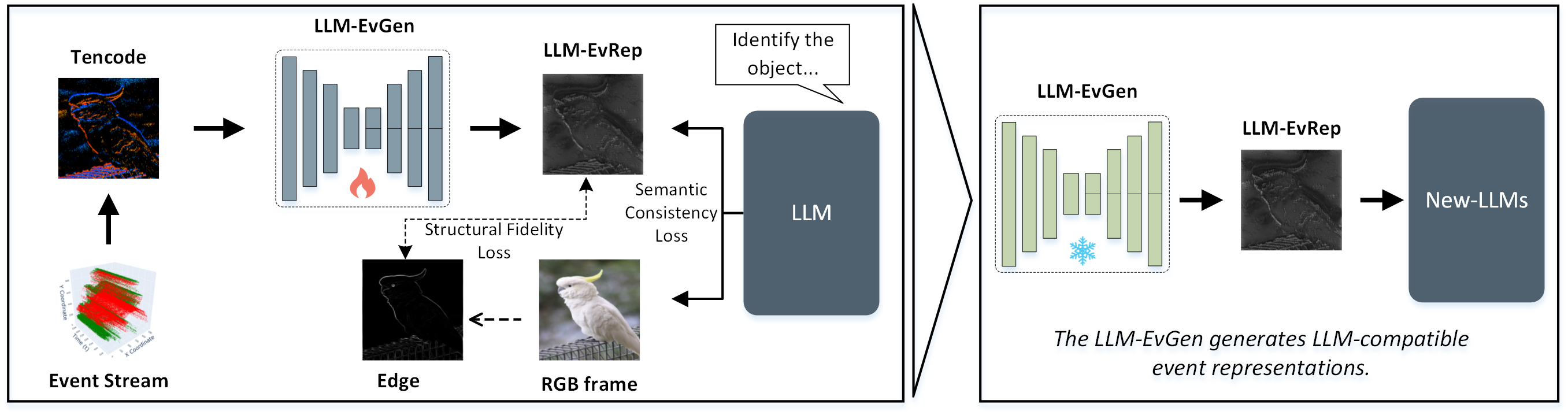}
  \caption{Overview of the proposed self-supervised learning framework for training an LLM-agnostic representation generator, \textbf{LLM-EvGen}, which generates an LLM-compatible event-stream representation, \textbf{LLM-EvRep}.
    On the left, the raw event stream is first transformed into a 3-channel Tencode representation. This representation is then used to train \textbf{LLM-EvGen} through two key loss functions: a structural fidelity loss derived from the RGB frame and a semantic consistency loss guided by the LLM.
    On the right, the trained \textbf{LLM-EvGen} can be seamlessly applied to other LLMs without requiring additional fine-tuning. It produces \textbf{LLM-EvRep}, which enhances the performance of LLMs in event-based zero-shot object recognition tasks.}
  \Description{Diagram illustrating the proposed self-supervised learning framework, LLM-EvGen, which generates LLM-compatible event-stream representations (LLM-EvRep). The left side shows the process of transforming a raw event stream into a 3-channel Tencode representation. LLM-EvGen is then trained using structural fidelity loss (from an RGB frame) and semantic consistency loss (guided by an LLM). On the right, the trained LLM-EvGen is applied to other LLMs without additional fine-tuning, improving event-based zero-shot object recognition.}
  \label{fig:teaser}
\end{teaserfigure}


\maketitle

\section{Introduction}
Event cameras, as bio-inspired sensors~\cite{gallego2020event}, have garnered significant attention in the computer vision community for their exceptional capabilities, including microsecond-level temporal resolution, high dynamic range (typically 140 dB compared to 60 dB of standard cameras), and low power consumption~\cite{gehrig2022high,son20174}. These advantages represent a paradigm shift from traditional frame-based imaging~\cite{gallego2020event,kong2024openess}, offering a novel approach to capturing visual information. Despite being in the early stages of development, event-driven visual perception has rapidly emerged as a critical area of research in contemporary computer vision~\cite{zheng2023deep}. Event-based vision has shown promising results across various applications, including object recognition~\cite{zheng2024eventdance,zhou2024eventbind}, semantic segmentation~\cite{kong2024openess,jia2023event}, detection~\cite{gehrig2024low,gehrig2023recurrent}, tracking~\cite{gallego2017event,gehrig2018asynchronous}, and optical flow estimation~\cite{lee2020spike,hagenaars2021self}.

Despite these advantages, leveraging event cameras for high-level visual tasks, such as object recognition, remains challenging. Existing approaches to event-based object recognition can be broadly categorized into two types: traditional neural network-based methods~\cite{zheng2024eventdance,su2023event} and CLIP-based zero-shot methods~\cite{wu2023eventclip,zhou2023clip,zhou2024eventbind}. Traditional neural network methods require extensive training, and due to the inherent limitations of neural networks, they are constrained to recognizing a limited set of categories~\cite{gehrig2024low}. To address these limitations, zero-shot methods~\cite{wu2023eventclip,zhou2024eventbind,zhou2023clip} have been proposed. While these approaches show promise in zero-shot open-world event-based object recognition, their reliance on CLIP introduces inherent limitations~\cite{yu2024can}. In contrast, large language models (LLMs) offer a compelling alternative, with richer pre-trained knowledge and superior zero-shot reasoning capabilities.

LLMs, with their exceptional zero-shot reasoning capabilities and vast pre-trained knowledge, have demonstrated remarkable success in multimodal tasks such as vision-language understanding and scene reasoning~\cite{zhu2023minigpt,fu2024scene}. Their ability to understand and generalize across diverse data modalities makes them a promising alternative to existing methods like CLIP for object recognition tasks. While LLMs have shown exceptional performance on 2D image-based content understanding, event-based visual content poses unique challenges due to its sparse, asynchronous, and modality-specific nature. These stark differences between event streams and traditional image-based inputs hinder LLMs' ability to directly comprehend event data. 

A critical challenge in applying LLMs to event-based vision is converting event streams into representations compatible with LLMs. Existing approaches to bridging this gap typically reconstruct event streams into more interpretable formats, falling into two main categories: event frame generation and event-to-video (E2V) techniques. The first category involves integrating events based on their spatial positions to generate ``event frames.'' The second category uses E2V methods, like E2VID~\cite{rebecq2019high} and E2HQV~\cite{qu2024e2hqv}, to reconstruct events into natural images, known as ``reconstructed frames.'' While these methods show promise in LLM-based zero-shot object recognition~\cite{yu2024can}, they do not specifically tailor event data to the unique characteristics of LLMs.

To address the limitations of existing methods and adapt event streams to LLMs, we propose \textbf{LLM-EvGen}, an event representation generator designed for producing LLM-compatible event representations. Inspired by E2HQV~\cite{qu2024e2hqv}, \textbf{LLM-EvGen} employs an encoder-decoder structure based on the MBConv and Fused MBConv layers from EfficientNetV2~\cite{tan2021efficientnetv2}, balancing computational efficiency and parameter utilization. The encoder-decoder structure provides a solid foundation for processing event data, but ensuring compatibility with LLMs requires careful supervision.As illustrated in Figure~\ref{fig:teaser}, we introduce a self-supervised learning framework that aligns event representations produced by \textbf{LLM-EvGen} with RGB frames, ensuring both semantic consistency and structural fidelity. Specifically, the event representations and their corresponding RGB frames are input into an LLM to extract semantic information, with consistency measured using Jaccard similarity~\cite{jaccard1901} as the \textbf{Semantic Consistency Loss}. To address early-stage noise in training, we propose a \textbf{Structural Fidelity Loss}, calculated as the mean squared error (MSE) between the Sobel edge maps~\cite{sobel1968} of \textbf{LLM-EvRep} and its RGB frames, enforcing structural consistency. \textbf{LLM-EvGen} requires training with only a single LLM and significantly improves event-based zero-shot object recognition performance. Extensive experiments validate the superiority of our approach in generating high-quality, LLM-compatible event representations and advancing event-based recognition tasks.

To sum up, our key contributions are as follows:
\begin{itemize}
    \item To the best of our knowledge, we are the first to explore how to enhance large LLMs' understanding of event-based visual content, bridging the gap between event-based data and LLMs.
    
    \item We propose \textbf{LLM-EvGen}, a novel event representation generator designed to produce LLM-compatible event representations, termed \textbf{LLM-EvRep}, which significantly improve LLM performance on event recognition tasks.
    
    \item We introduce a self-supervised learning framework that combines LLM-driven \textbf{Semantic Consistency Loss} and an auxiliary \textbf{Structural Fidelity Loss} to effectively train \textbf{LLM-EvGen}, ensuring both semantic alignment and structural consistency in the generated representations.
    
    \item Extensive experimental results demonstrate that our \textbf{LLM-EvRep} achieves higher recognition accuracy than hand-crafted and E2V methods across multiple benchmark datasets, setting a new benchmark for leveraging Large Language Models (LLMs) in event-based vision tasks.

\end{itemize}

\section{Related Work}
\noindent \textbf{LLMs for Visual Understanding.} Classical LLMs such as OpenAI's GPT-4/4 Turbo~\cite{openai2024gpt4}, Meta's LLaMA 3.1~\cite{dubey2024llama}, LLaVA~\cite{liu2024improved}, and MiniGPT-4-v2~\cite{chen2023minigptv2}, trained on vast amounts of cross-modal data, can be directly applied to various downstream tasks, such as object recognition and visual question answering, without the need for additional training or fine-tuning. Zang et al.~\cite{zang2024contextual} proposed ContextDET, a framework combining a visual encoder, pre-trained LLM, and visual decoder for context-aware object detection in human-AI interactions. Lv et al.~\cite{tang2024chain} introduced the Multimodal Camo-Perceptive Framework (MMCPF), using a Chain of Visual Perception strategy to improve zero-shot Camouflaged Object Detection (COD) with LLMs. Zhu et al.~\cite{zhu2024multi} developed a depth-aware Transformer to integrate object depth information for improved Visual Commonsense Reasoning (VCR) by considering 3D spatial relationships in visual and textual data.
Despite their impressive multimodal capabilities, LLMs face challenges in understanding event-driven visual content due to the substantial modality differences from traditional images. In this work, we leverage the superior cross-modal understanding capabilities of LLMs to train an LLM-compatible event representation generator. This generator effectively bridges the modality gap between events and RGB frames, thereby enhancing the ability of LLMs to comprehend event-based visual content.

\begin{figure}[t]
  \centering
  \includegraphics[width=0.35\textwidth]{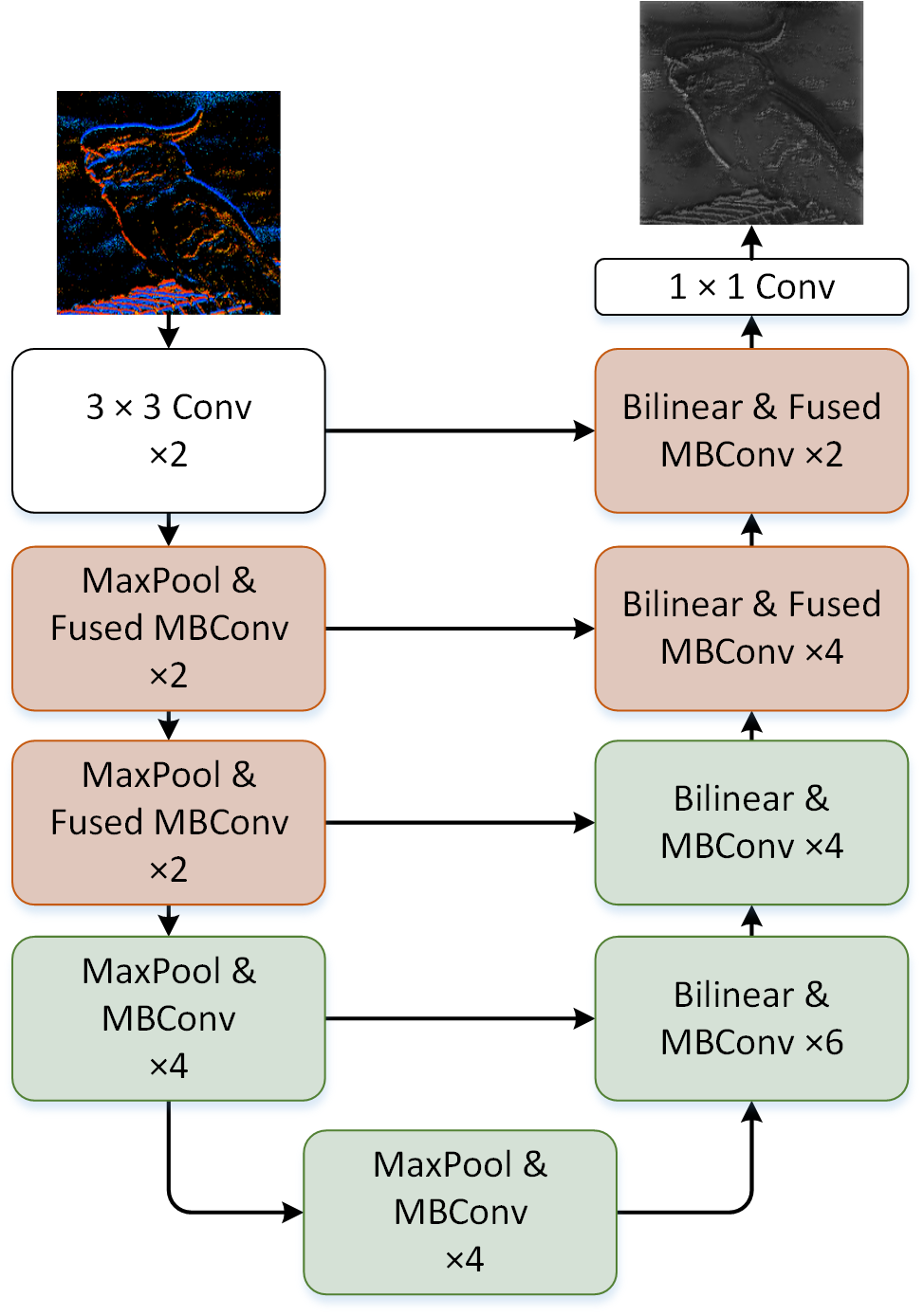}
  \caption{Structure of the Proposed LLM-EvGen. An encoder-decoder architecture consisting of MBConv and Fused MBConv~\cite{tan2021efficientnetv2} layers.}
  \Description{Diagram illustrating the structure of the proposed LLM-EvGen, an encoder-decoder architecture. The network begins with a raw event stream input, which is first processed through two 3×3 convolution layers. It then passes through multiple stages, including max pooling, and a series of MBConv and Fused MBConv layers. The decoder follows a bilinear upsampling approach, also incorporating MBConv and Fused MBConv layers. The final output is generated using a 1×1 convolution.}
  \label{fig:network framework}
\end{figure}

\noindent \textbf{Event-based Object Recognition.} Event cameras, characterized by their low latency, high dynamic range, and low power consumption~\cite{gehrig2022high,son20174}, have gained prominence in computer vision applications and have shown promising results in object recognition. Existing object recognition approaches can be broadly classified into traditional neural network-based methods~\cite{zheng2024eventdance,su2023event} and CLIP-based zero-shot methods~\cite{ wu2023eventclip,zhou2023clip,zhou2024eventbind}. For traditional neural network-based methods, learning high-performance models for event data presents significant challenges due to the asynchronous nature of events and the limited availability of large-scale labeled datasets~\cite{wu2023eventclip}. Moreover, these methods struggle with recognizing new event categories, as retraining large models for every new category is computationally expensive and impractical~\cite{zhou2024eventbind,gehrig2024low}. To address these challenges, researchers have proposed CLIP-based zero-shot methods, leveraging CLIP’s extensive pre-trained knowledge to bridge the modality gap between events, images, and texts. While these methods mitigate some limitations of traditional approaches, they remain inherently constrained by the characteristics of CLIP~\cite{yu2024can}. In contrast to existing methods, this paper explores the potential of LLMs with their extensive pre-trained knowledge and advanced multi-modal understanding capabilities. We aim to train an event generator in a self-supervised manner by distilling knowledge from pre-trained LLMs. This approach seeks to overcome the limitations of scalability and the dependence on large-scale annotated datasets, addressing key challenges in event-based visual recognition.

\section{Method}
In this section, we provide a detailed introduction to the framework for generating LLM-compatible event stream representations through self-supervised learning. The process starts by converting the raw event stream into a neural network-friendly representation, Tencode~\cite{huang2023eventpoint}. This representation is then fed into our LLM-compatible Event Representation Generator, \textbf{LLM-EvGen}, an encoder-decoder network inspired by E2HQV~\cite{qu2024e2hqv} and EfficientNetV2~\cite{tan2021efficientnetv2} (see Section~\ref{section3.1}). Subsequently, \textbf{LLM-EvGen} is trained using a self-supervised framework based on semantic consistency, enabling it to produce LLM-compatible event representations, referred to as \textbf{LLM-EvRep} (see Section~\ref{section3.2}). To accelerate model convergence and enhance the quality of representations, an additional structural fidelity loss is introduced during training. This ensures that the generated representations not only retain structural consistency but also adapt effectively to LLMs, thereby facilitating improved performance in event-based recognition tasks performed by LLMs.

\subsection{Learnable Event Representation Generator}
\label{section3.1}
Inspired by E2HQV~\cite{qu2024e2hqv}, the proposed \textbf{LLM-EvGen} adopts a U-Net-like architecture, as shown in Figure~\ref{fig:network framework}. Following this design, \textbf{LLM-EvGen} incorporates the MBConv and Fused MBConv layers from EfficientNetV2~\cite{tan2021efficientnetv2}. These layers are chosen for their optimal balance between training efficiency and parameter utilization~\cite{qu2024e2hqv}, making it well-suited for our model.

Before feeding the discretized event stream into \textbf{LLM-EvGen}, we first convert it into a neural network-compatible representation. Following Huang et al.~\cite{huang2023eventpoint}, we transform the event stream into a structured frame representation, denoted as $F$, as shown in Figure~\ref{fig:teaser}. Then the event stream representation $F$ is processed by the encoder of the proposed \textbf{LLM-EvGen}, which extracts hierarchical features while progressively reducing the spatial resolution. The encoder begins with two $3\times3$ convolutional layers to capture low-level features and improve spatial representations. Subsequently, the network incorporates a sequence of MaxPooling layers combined with Fused MBConv and MBConv layers, inspired by EfficientNetV2~\cite{tan2021efficientnetv2} and E2HQV~\cite{qu2024e2hqv}. These layers effectively balance computational efficiency and feature extraction capability, enabling the encoder to capture both local and global features.

As the feature maps flow through the network, the spatial resolution is progressively reduced using MaxPooling, while the depth of the features is increased via MBConv layers with varying numbers of repetitions. This design allows the network to extract complex representations at multiple scales, facilitating robust feature learning from the event data.

In the decoder, the feature maps are gradually upsampled using bilinear interpolation, followed by MBConv and Fused MBConv layers to refine the features and recover spatial resolution. Skip connections between the encoder and decoder ensure that low-level spatial details are preserved and fused with high-level semantic information during reconstruction. Finally, $1\times1$ convolutional layers are employed to reduce the channel dimensions and generate the final output representation.
This hierarchical encoding-decoding architecture enables \textbf{LLM-EvGen} to effectively process the event-based input $F$, capturing both the fine-grained details and high-level semantics necessary for the downstream tasks.

\subsection{Self-supervised Learning Framework}
\label{section3.2}
To make \textbf{LLM-EvGen} compatible with LLMs, we propose a self-supervised framework based on semantic consistency . Specifically, as illustrated in Figure~\ref{fig:teaser}, after processing $F$ with \textbf{LLM-EvGen}, the model generates corresponding \textbf{LLM-EvRep}. To extract semantic information from \textbf{LLM-EvRep}, we directly input it into the LLM (in our case, LLaVA~\cite{liu2023improvedllava}) and obtain its semantic information. Simultaneously, the corresponding RGB frame (denoted as RGB) is also fed into the LLM to generate a comparable semantic information, this processing can be describe as following:

\begin{equation}
\begin{aligned}
f^{e} = \text{LLM}(\text{LLM-EvRep}) \\
f^{r} = \text{LLM}(\text{RGB})
\end{aligned}
\end{equation}

\noindent where $f^{e}$ represents semantic information of the \textbf{LLM-EvRep} and $f^{r}$ is semantic information of its corresponding RGB frame.

To ensure semantic alignment between \textbf{LLM-EvRep} and their corresponding RGB frame, we define a \textbf{semantic consistency loss} based on the semantic information generated by the LLM. Specifically, the semantic information $f^e$ and $f^r$ are the textual outputs produced by the LLM when processing \textbf{LLM-EvRep} and the corresponding RGB frame, respectively. Our goal is to minimize the semantic discrepancy between these two pieces of semantic information.

To achieve this, we compute a Jaccard similarity-based~\cite{jaccard1901} \textbf{Semantic Consistency Loss}. First, $f^e$ and $f^r$ are tokenized into sets of words, denoted as $W_e$ and $W_r$, respectively:

\begin{equation}
\begin{aligned}
W_e = \text{set}(\text{tokenize}(f^e)), \quad W_r = \text{set}(\text{tokenize}(f^r)).
\end{aligned}
\end{equation}

The semantic similarity between these two sets is measured using the Jaccard similarity, which is defined as the \textbf{Semantic Consistency Loss}. This loss quantifies the dissimilarity between the textual representations generated by the LLM from \textbf{LLM-EvRep} and its corresponding RGB-based inputs:

\begin{equation} L_{\text{semantic}} = 1 - \frac{|W_e \cap W_r|}{|W_e \cup W_r|}. \end{equation}

\noindent where $|W_e \cap W_r|$ and $|W_e \cup W_r|$ represent the sizes of the intersection and union of $W_e$ and $W_r$, respectively. This loss penalizes semantic differences, encouraging \textbf{LLM-EvGen} to produce event representations that are semantically consistent with their corresponding RGB frames.

By aligning the semantic consistency between \textbf{LLM-EvRep} and the corresponding RGB frame, the proposed framework effectively guides \textbf{LLM-EvGen} to generate event representations that are semantically aligned to their RGB counterparts. 

However, due to the U-Net-like architecture of \textbf{LLM-EvGen}, the event representations generated during the early stages of training often contain significant noise. This noise can hinder the LLM's ability to extract meaningful semantic information from these representations. To address this issue, we introduce an auxiliary weak supervision mechanism that enforces structural consistency between the event-based representations and their corresponding RGB frames.

Specifically, as presented in Figure~\ref{fig:teaser}, we design an auxiliary \textbf{Structural Fidelity Loss}, which leverages sobel edge detection~\cite{sobel1968} to preserve spatial and structural details in the generated event-based frames. The sobel edge detection method computes the gradient magnitude of an image in both the horizontal and vertical directions using predefined sobel kernels. For an input image $\mathbf{I}$, the gradients in the $x$- and $y$-directions, denoted as $G_x$ and $G_y$, are computed as:

\begin{equation}
\begin{aligned}
G_x = \mathbf{I} * K_x, \quad G_y = \mathbf{I} * K_y,
\end{aligned}
\end{equation}

\noindent where $K_x$ and $K_y$ are the sobel kernels for the horizontal and vertical directions, respectively, and $*$ denotes the convolution operation. The gradient magnitude is then computed as:

\begin{equation}
\begin{aligned}
G = \sqrt{G_x^2 + G_y^2}.
\end{aligned}
\end{equation}

This gradient magnitude map highlights the edges in the image, capturing the structural information present in the input. To ensure consistency between the structural details of the generated \textbf{LLM-EvRep} and its corresponding RGB frames, we define the \textbf{Structural Fidelity Loss} as the mean squared error (MSE) between their sobel edge map:

\begin{equation}
\begin{aligned}
L_{\text{fidelity}} = \text{MSE}(\text{sobel}(\mathbf{O}), \text{sobel}(\mathbf{T})),
\end{aligned}
\end{equation}

\noindent where $\mathbf{O}$ and $\mathbf{T}$ represent the sobel edge maps extracted from \textbf{LLM-EvRep} and its corresponding RGB frame, respectively, and $\text{sobel}(\cdot)$ computes the sobel edge map of the input.

To jointly optimize the semantic alignment and structural consistency of the \textbf{LLM-EvRep}, we combine the \textbf{Semantic Consistency Loss} and the \textbf{Structural Fidelity Loss} into a unified objective, referred to as the \textbf{Dual Alignment Loss}:

\begin{equation}
\begin{aligned}
L_{\text{dual}} = \lambda L_{\text{semantic}} + \gamma L_{\text{fidelity}},
\end{aligned}
\end{equation}

\noindent where $\lambda$ and $\gamma$ are weighting factors that balance the contributions of the two losses. The \textbf{Semantic Consistency Loss} ensures that \textbf{LLM-EvRep} generated by \textbf{LLM-EvGen} are semantically aligned with their RGB counterparts, while the \textbf{Structural Fidelity Loss} enforces spatial and structural consistency in the representations.

This dual-loss strategy integrates semantic alignment and structural precision, effectively mitigating noise during the early stages of training. By doing so, it enhances the robustness of the learned representations and accelerates model convergence. The \textbf{Dual Alignment Loss} ensures that \textbf{LLM-EvGen} produces high-quality event representations that are both semantically meaningful and structurally precise.


\begin{table}[h]
\centering
\caption{Comparison of Classification Accuracy Across N-ImageNet (N-I), N-Caltech101 (N-C), and N-MNIST (N-M) Datasets. Due to the prompt length limitations of LLaVA and MiniGPT-4-v2, we are unable to evaluate their recognition performance on the N-ImageNet dataset. \textbf{Bold values indicate the best performance for each dataset.}}
\begin{tabular}{ccccc}
\hline
\multirow{2}{*}{Method} & \multirow{2}{*}{Input} & \multicolumn{3}{c}{Accuracy (\%)} \\ \cline{3-5} 
                        &                             & N-I  & N-C  & N-M  \\ \hline
ECLIP                   & Event frame                       & 8.72        & 53.88         & 14.56      \\
EventCLIP               & Event stream                       & 4.30        & 49.95         & 11.87      \\ 
EventBind               & Event frame                      & 3.05        & 67.58         & 12.89      \\  \hline 
\multirow{4}{*}{\begin{tabular}[c]{@{}c@{}}MiniGPT-4-v2\textsuperscript{*}\end{tabular}} 
                        & Event frame                         & -           & 23.10         & 15.29      \\ 
                        & E2VID                       & -           & 30.20         & 10.70      \\ 
                        & E2HQV                       & -           & 30.71         & 11.62      \\ 
                        & LLM-EvRep (ours)            & -           & \textbf{33.68}             & \textbf{17.53}          \\ \hline
\multirow{4}{*}{\begin{tabular}[c]{@{}c@{}}LLaVA\textsuperscript{*}\end{tabular}}    
                        & Event frame                         & -           & 64.72         & 56.27      \\ 
                        & E2VID                       & -           & 67.26         & 19.57      \\ 
                        & E2HQV                       & -           & 67.51         & 22.94      \\ 
                        & LLM-EvRep (ours)            & -           & \textbf{69.94}             & \textbf{88.07}          \\ \hline
\multirow{4}{*}{GPT-4turbo} 
                        & Event frame                         & 17.23       & 72.08         & 85.95      \\ 
                        & E2VID                       & 11.61       & 55.08         & 39.14      \\ 
                        & E2HQV                       & 10.70       & 55.58         & 39.41      \\ 
                        & LLM-EvRep (ours)            & \textbf{20.02}           & \textbf{75.32}             & \textbf{92.39}          \\ \hline
\multirow{4}{*}{GPT-4o} 
                        & Event frame                         & 45.55       & 89.37         & 91.86      \\ 
                        & E2VID                       & 32.75       & 93.90 & 49.79      \\ 
                        & E2HQV                       & 30.05       & 93.23         & 51.68      \\ 
                        & LLM-EvRep (ours)            & \textbf{48.68}           & \textbf{94.72}             & \textbf{100}          \\ \hline
\end{tabular}
\label{tab:comparison_accuracy}
\footnotesize \textsuperscript{*}Open-source model
\end{table}

\section{Experimental}
\subsection{Setup}
\noindent \textbf{Evaluation LLMs.}
We evaluated the recognition accuracy of various event representations across two open-source models, LLaVA~\cite{liu2023improvedllava} and MiniGPT-4-v2~\cite{chen2023minigptv2}, and two proprietary models, GPT-4o and GPT-4 Turbo~\cite{openai2024gpt4}. Furthermore, we compared the performance of three CLIP-based zero-shot methods: ECLIP~\cite{zhou2023clip}, EventCLIP~\cite{wu2023eventclip}, and EventBind~\cite{zhou2024eventbind}.

\noindent \textbf{Training datasets.} We use the N-ImageNet~\cite{kim2021n} dataset to train our \textbf{LLM-EvGen}. N-ImageNet is derived from the ImageNet-1K~\cite{deng2009imagenet} dataset, where RGB images are displayed on a monitor and captured by a moving event camera. The dataset consists of 1,781,167 event streams with a resolution of $480\times640$, covering 1,000 unique object classes. We split the dataset into training and testing sets with a ratio of 5:1, where the testing set is used for evaluation experiments.

\noindent \textbf{Downstream tasks datasets.} To evaluate the performance of \textbf{LLM-EvRep}, we conduct experiments on object recognition task. We utilize the N-ImageNet~\cite{kim2021n}, N-Caltech101~\cite{fei2004learning}, and N-MNIST~\cite{orchard2015converting} datasets. 

\noindent \textbf{Implementation Details.}  
We utilize LLaVA~\cite{liu2024improved} to train our \textbf{LLM-EvGen}, with the weights of LLaVA frozen throughout the training process. The \textbf{LLM-EvGen} is trained for 50 epochs with a batch size of 16 and a learning rate of $1\mathrm{e}{-4}$, using the Adam optimizer~\cite{kingma2014adam}. All experiments are conducted on an NVIDIA RTX 3090 GPU.

\subsection{Results}
Table~\ref{tab:comparison_accuracy} presents the quantitative evaluation results. As shown, compared to CLIP-based zero-shot methods, our proposed \textbf{LLM-EvRep} achieves an accuracy of 94.72\% on the N-Caltech101 dataset using GPT-4o, surpassing the state-of-the-art zero-shot method EventBind~\cite{zhou2024eventbind} by 27.14\%. On the N-ImageNet dataset, our method outperforms ECLIP~\cite{zhou2023clip} by nearly six orders of magnitude, reaching an accuracy of almost 50\%. These results clearly demonstrate that our LLM-based approach exhibits superior performance compared to traditional CLIP-based zero-shot methods in event-driven vision tasks. 

Further examining Table~\ref{tab:comparison_accuracy}, we used the representative open-source models LLaVA~\cite{liu2024improved} and MiniGPT-4-v2~\cite{chen2023minigptv2} for recognition tasks. The results show that our proposed \textbf{LLM-EvRep} method outperforms both hand-crafted and E2V methods in recognition accuracy. Specifically, experiments on the N-Caltech101 dataset show that \textbf{LLM-EvRep} achieves an accuracy improvement of 5.22\% over the hand-crafted event frames, and surpasses the V2E methods, E2VID and E2HQV, by 2.68\% and 2.43\%, respectively. Furthermore, results on the N-MNIST dataset indicate that \textbf{LLM-EvRep} improves accuracy by 31.80\% over the event frames and is approximately four times more accurate than E2VID and E2HQV. These results demonstrate that our method effectively enhances the image understanding ability of open-source LLMs.

\begin{table}[]
\caption{Comparison of accuracy between the proposed LLM-EvRep and Tencode representations on the N-Caltech101 and N-MNIST datasets, evaluated using LLaVA and GPT-4o.}
\begin{tabular}{c|c|c|c}
\hline
       & Input     & N-Caltech101 & N-MNIST \\ \hline
\multirow{2}{*}{LLaVA\textsuperscript{*}}  
       & Tencode   & 68.18           & 66.97 \\ \cline{2-4}
       & LLM-EvRep & \textbf{69.94}        & \textbf{88.07 }\\ \hline
\multirow{2}{*}{GPT-4o} 
       & Tencode   & 89.34            & 99.38 \\ \cline{2-4}
       & LLM-EvRep & \textbf{94.72}        & \textbf{100} \\ \hline
\end{tabular}
\label{tab:comparison_with_Tencode}
\end{table}

At the same time, we also conducted recognition experiments on the representative proprietary models GPT-4o and GPT-4uurbo. The experimental results show that \textbf{LLM-EvRep} outperforms both the hand-crafted and E2V methods across the three experimental datasets. Notably, \textbf{LLM-EvRep} achieved an astonishing 100\% recognition accuracy on the N-MNIST dataset, which is nearly twice as high as the two E2V methods. This strongly demonstrates the superior performance of our method on simpler datasets.

Meanwhile, we also compare the accuracy with Tencode~\cite{huang2023eventpoint}, as shown in Table~\ref{tab:comparison_with_Tencode}. The results in Table~\ref{tab:comparison_with_Tencode} indicate that, whether evaluated on the open-source model LLaVA or the proprietary model GPT-4o, our proposed \textbf{LLM-EvRep} consistently outperforms Tencode in terms of accuracy. These findings further validate the effectiveness of our approach.

In summary, the experimental results across multiple datasets and models consistently demonstrate the superior performance of our proposed \textbf{LLM-EvRep} method in event-driven vision tasks. Compared to existing CLIP-based zero-shot methods, hand-crafted approaches, and state-of-the-art E2V techniques, our method achieves higher recognition accuracy. These findings highlight the potential of integrating LLMs with event stream representations, offering a promising direction for advancing the capabilities of event-driven visual tasks.

\section{Conclusion}
In this paper, we introduce \textbf{LLM-EvGen}, a novel event representation generator specifically designed for large language models (LLMs). To effectively train LLM-EvGen, we propose a self-supervised learning framework based on a dual alignment approach that ensures both \textbf{LLM Semantic Consistency} and \textbf{Structural Fidelity}. This framework leverages two key components: a semantic alignment loss that ensures the generated event representations are semantically coherent with the corresponding RGB frames, and a structural fidelity loss that preserves the spatial integrity of the event data. Extensive experimental evaluations demonstrate the effectiveness of our proposed approach.



\bibliographystyle{ACM-Reference-Format}
\bibliography{sample-base}


\begin{thebibliography}{41}


\ifx \showCODEN    \undefined \def \showCODEN     #1{\unskip}     \fi
\ifx \showISBNx    \undefined \def \showISBNx     #1{\unskip}     \fi
\ifx \showISBNxiii \undefined \def \showISBNxiii  #1{\unskip}     \fi
\ifx \showISSN     \undefined \def \showISSN      #1{\unskip}     \fi
\ifx \showLCCN     \undefined \def \showLCCN      #1{\unskip}     \fi
\ifx \shownote     \undefined \def \shownote      #1{#1}          \fi
\ifx \showarticletitle \undefined \def \showarticletitle #1{#1}   \fi
\ifx \showURL      \undefined \def \showURL       {\relax}        \fi
\providecommand\bibfield[2]{#2}
\providecommand\bibinfo[2]{#2}
\providecommand\natexlab[1]{#1}
\providecommand\showeprint[2][]{arXiv:#2}

\bibitem[Chen et~al\mbox{.}(2023)]%
        {chen2023minigptv2}
\bibfield{author}{\bibinfo{person}{Jun Chen}, \bibinfo{person}{Deyao Zhu}, \bibinfo{person}{Xiaoqian Shen}, \bibinfo{person}{Xiang Li}, \bibinfo{person}{Zechun Liu}, \bibinfo{person}{Pengchuan Zhang}, \bibinfo{person}{Raghuraman Krishnamoorthi}, \bibinfo{person}{Vikas Chandra}, \bibinfo{person}{Yunyang Xiong}, {and} \bibinfo{person}{Mohamed Elhoseiny}.} \bibinfo{year}{2023}\natexlab{}.
\newblock \showarticletitle{MiniGPT-v2: Large Language Model as a Unified Interface for Vision-Language Multi-Task Learning}.
\newblock \bibinfo{journal}{\emph{arXiv preprint arXiv:2310.09478}} (\bibinfo{year}{2023}).
\newblock


\bibitem[Deng et~al\mbox{.}(2009)]%
        {deng2009imagenet}
\bibfield{author}{\bibinfo{person}{Jia Deng}, \bibinfo{person}{Wei Dong}, \bibinfo{person}{Richard Socher}, \bibinfo{person}{Li-Jia Li}, \bibinfo{person}{Kai Li}, {and} \bibinfo{person}{Li Fei-Fei}.} \bibinfo{year}{2009}\natexlab{}.
\newblock \showarticletitle{Imagenet: A large-scale hierarchical image database}. In \bibinfo{booktitle}{\emph{2009 IEEE conference on computer vision and pattern recognition}}. Ieee, \bibinfo{pages}{248--255}.
\newblock


\bibitem[Dubey et~al\mbox{.}(2024)]%
        {dubey2024llama}
\bibfield{author}{\bibinfo{person}{Abhimanyu Dubey}, \bibinfo{person}{Abhinav Jauhri}, \bibinfo{person}{Abhinav Pandey}, \bibinfo{person}{Abhishek Kadian}, \bibinfo{person}{Ahmad Al-Dahle}, \bibinfo{person}{Aiesha Letman}, \bibinfo{person}{Akhil Mathur}, \bibinfo{person}{Alan Schelten}, \bibinfo{person}{Amy Yang}, \bibinfo{person}{Angela Fan}, {et~al\mbox{.}}} \bibinfo{year}{2024}\natexlab{}.
\newblock \showarticletitle{The llama 3 herd of models}.
\newblock \bibinfo{journal}{\emph{arXiv preprint arXiv:2407.21783}} (\bibinfo{year}{2024}).
\newblock


\bibitem[Fei-Fei et~al\mbox{.}(2004)]%
        {fei2004learning}
\bibfield{author}{\bibinfo{person}{Li Fei-Fei}, \bibinfo{person}{Rob Fergus}, {and} \bibinfo{person}{Pietro Perona}.} \bibinfo{year}{2004}\natexlab{}.
\newblock \showarticletitle{Learning generative visual models from few training examples: An incremental bayesian approach tested on 101 object categories}. In \bibinfo{booktitle}{\emph{2004 conference on computer vision and pattern recognition workshop}}. IEEE, \bibinfo{pages}{178--178}.
\newblock


\bibitem[Fu et~al\mbox{.}(2024)]%
        {fu2024scene}
\bibfield{author}{\bibinfo{person}{Rao Fu}, \bibinfo{person}{Jingyu Liu}, \bibinfo{person}{Xilun Chen}, \bibinfo{person}{Yixin Nie}, {and} \bibinfo{person}{Wenhan Xiong}.} \bibinfo{year}{2024}\natexlab{}.
\newblock \showarticletitle{Scene-llm: Extending language model for 3d visual understanding and reasoning}.
\newblock \bibinfo{journal}{\emph{arXiv preprint arXiv:2403.11401}} (\bibinfo{year}{2024}).
\newblock


\bibitem[Gallego et~al\mbox{.}(2020)]%
        {gallego2020event}
\bibfield{author}{\bibinfo{person}{Guillermo Gallego}, \bibinfo{person}{Tobi Delbr{\"u}ck}, \bibinfo{person}{Garrick Orchard}, \bibinfo{person}{Chiara Bartolozzi}, \bibinfo{person}{Brian Taba}, \bibinfo{person}{Andrea Censi}, \bibinfo{person}{Stefan Leutenegger}, \bibinfo{person}{Andrew~J Davison}, \bibinfo{person}{J{\"o}rg Conradt}, \bibinfo{person}{Kostas Daniilidis}, {et~al\mbox{.}}} \bibinfo{year}{2020}\natexlab{}.
\newblock \showarticletitle{Event-based vision: A survey}.
\newblock \bibinfo{journal}{\emph{IEEE transactions on pattern analysis and machine intelligence}} \bibinfo{volume}{44}, \bibinfo{number}{1} (\bibinfo{year}{2020}), \bibinfo{pages}{154--180}.
\newblock


\bibitem[Gallego et~al\mbox{.}(2017)]%
        {gallego2017event}
\bibfield{author}{\bibinfo{person}{Guillermo Gallego}, \bibinfo{person}{Jon~EA Lund}, \bibinfo{person}{Elias Mueggler}, \bibinfo{person}{Henri Rebecq}, \bibinfo{person}{Tobi Delbruck}, {and} \bibinfo{person}{Davide Scaramuzza}.} \bibinfo{year}{2017}\natexlab{}.
\newblock \showarticletitle{Event-based, 6-DOF camera tracking from photometric depth maps}.
\newblock \bibinfo{journal}{\emph{IEEE transactions on pattern analysis and machine intelligence}} \bibinfo{volume}{40}, \bibinfo{number}{10} (\bibinfo{year}{2017}), \bibinfo{pages}{2402--2412}.
\newblock


\bibitem[Gehrig et~al\mbox{.}(2018)]%
        {gehrig2018asynchronous}
\bibfield{author}{\bibinfo{person}{Daniel Gehrig}, \bibinfo{person}{Henri Rebecq}, \bibinfo{person}{Guillermo Gallego}, {and} \bibinfo{person}{Davide Scaramuzza}.} \bibinfo{year}{2018}\natexlab{}.
\newblock \showarticletitle{Asynchronous, photometric feature tracking using events and frames}. In \bibinfo{booktitle}{\emph{Proceedings of the European Conference on Computer Vision (ECCV)}}. \bibinfo{pages}{750--765}.
\newblock


\bibitem[Gehrig and Scaramuzza(2022)]%
        {gehrig2022high}
\bibfield{author}{\bibinfo{person}{Daniel Gehrig} {and} \bibinfo{person}{Davide Scaramuzza}.} \bibinfo{year}{2022}\natexlab{}.
\newblock \showarticletitle{Are high-resolution event cameras really needed?}
\newblock \bibinfo{journal}{\emph{arXiv preprint arXiv:2203.14672}} (\bibinfo{year}{2022}).
\newblock


\bibitem[Gehrig and Scaramuzza(2024)]%
        {gehrig2024low}
\bibfield{author}{\bibinfo{person}{Daniel Gehrig} {and} \bibinfo{person}{Davide Scaramuzza}.} \bibinfo{year}{2024}\natexlab{}.
\newblock \showarticletitle{Low-latency automotive vision with event cameras}.
\newblock \bibinfo{journal}{\emph{Nature}} \bibinfo{volume}{629}, \bibinfo{number}{8014} (\bibinfo{year}{2024}), \bibinfo{pages}{1034--1040}.
\newblock


\bibitem[Gehrig and Scaramuzza(2023)]%
        {gehrig2023recurrent}
\bibfield{author}{\bibinfo{person}{Mathias Gehrig} {and} \bibinfo{person}{Davide Scaramuzza}.} \bibinfo{year}{2023}\natexlab{}.
\newblock \showarticletitle{Recurrent vision transformers for object detection with event cameras}. In \bibinfo{booktitle}{\emph{Proceedings of the IEEE/CVF conference on computer vision and pattern recognition}}. \bibinfo{pages}{13884--13893}.
\newblock


\bibitem[Hagenaars et~al\mbox{.}(2021)]%
        {hagenaars2021self}
\bibfield{author}{\bibinfo{person}{Jesse Hagenaars}, \bibinfo{person}{Federico Paredes-Vall{\'e}s}, {and} \bibinfo{person}{Guido De~Croon}.} \bibinfo{year}{2021}\natexlab{}.
\newblock \showarticletitle{Self-supervised learning of event-based optical flow with spiking neural networks}.
\newblock \bibinfo{journal}{\emph{Advances in Neural Information Processing Systems}}  \bibinfo{volume}{34} (\bibinfo{year}{2021}), \bibinfo{pages}{7167--7179}.
\newblock


\bibitem[Huang et~al\mbox{.}(2023)]%
        {huang2023eventpoint}
\bibfield{author}{\bibinfo{person}{Ze Huang}, \bibinfo{person}{Li Sun}, \bibinfo{person}{Cheng Zhao}, \bibinfo{person}{Song Li}, {and} \bibinfo{person}{Songzhi Su}.} \bibinfo{year}{2023}\natexlab{}.
\newblock \showarticletitle{Eventpoint: Self-supervised interest point detection and description for event-based camera}. In \bibinfo{booktitle}{\emph{Proceedings of the IEEE/CVF Winter Conference on Applications of Computer Vision}}. \bibinfo{pages}{5396--5405}.
\newblock


\bibitem[Jaccard(1901)]%
        {jaccard1901}
\bibfield{author}{\bibinfo{person}{Paul Jaccard}.} \bibinfo{year}{1901}\natexlab{}.
\newblock \showarticletitle{Étude comparative de la distribution florale dans une portion des Alpes et des Jura}.
\newblock \bibinfo{journal}{\emph{Bulletin de la Société Vaudoise des Sciences Naturelles}}  \bibinfo{volume}{37} (\bibinfo{year}{1901}), \bibinfo{pages}{547--579}.
\newblock


\bibitem[Jia et~al\mbox{.}(2023)]%
        {jia2023event}
\bibfield{author}{\bibinfo{person}{Zexi Jia}, \bibinfo{person}{Kaichao You}, \bibinfo{person}{Weihua He}, \bibinfo{person}{Yang Tian}, \bibinfo{person}{Yongxiang Feng}, \bibinfo{person}{Yaoyuan Wang}, \bibinfo{person}{Xu Jia}, \bibinfo{person}{Yihang Lou}, \bibinfo{person}{Jingyi Zhang}, \bibinfo{person}{Guoqi Li}, {et~al\mbox{.}}} \bibinfo{year}{2023}\natexlab{}.
\newblock \showarticletitle{Event-based semantic segmentation with posterior attention}.
\newblock \bibinfo{journal}{\emph{IEEE Transactions on Image Processing}}  \bibinfo{volume}{32} (\bibinfo{year}{2023}), \bibinfo{pages}{1829--1842}.
\newblock


\bibitem[Kim et~al\mbox{.}(2021)]%
        {kim2021n}
\bibfield{author}{\bibinfo{person}{Junho Kim}, \bibinfo{person}{Jaehyeok Bae}, \bibinfo{person}{Gangin Park}, \bibinfo{person}{Dongsu Zhang}, {and} \bibinfo{person}{Young~Min Kim}.} \bibinfo{year}{2021}\natexlab{}.
\newblock \showarticletitle{N-imagenet: Towards robust, fine-grained object recognition with event cameras}. In \bibinfo{booktitle}{\emph{Proceedings of the IEEE/CVF international conference on computer vision}}. \bibinfo{pages}{2146--2156}.
\newblock


\bibitem[Kingma(2014)]%
        {kingma2014adam}
\bibfield{author}{\bibinfo{person}{Diederik~P Kingma}.} \bibinfo{year}{2014}\natexlab{}.
\newblock \showarticletitle{Adam: A method for stochastic optimization}.
\newblock \bibinfo{journal}{\emph{arXiv preprint arXiv:1412.6980}} (\bibinfo{year}{2014}).
\newblock


\bibitem[Kong et~al\mbox{.}(2024)]%
        {kong2024openess}
\bibfield{author}{\bibinfo{person}{Lingdong Kong}, \bibinfo{person}{Youquan Liu}, \bibinfo{person}{Lai~Xing Ng}, \bibinfo{person}{Benoit~R Cottereau}, {and} \bibinfo{person}{Wei~Tsang Ooi}.} \bibinfo{year}{2024}\natexlab{}.
\newblock \showarticletitle{OpenESS: Event-based Semantic Scene Understanding with Open Vocabularies}. In \bibinfo{booktitle}{\emph{Proceedings of the IEEE/CVF Conference on Computer Vision and Pattern Recognition}}. \bibinfo{pages}{15686--15698}.
\newblock


\bibitem[Lee et~al\mbox{.}(2020)]%
        {lee2020spike}
\bibfield{author}{\bibinfo{person}{Chankyu Lee}, \bibinfo{person}{Adarsh~Kumar Kosta}, \bibinfo{person}{Alex~Zihao Zhu}, \bibinfo{person}{Kenneth Chaney}, \bibinfo{person}{Kostas Daniilidis}, {and} \bibinfo{person}{Kaushik Roy}.} \bibinfo{year}{2020}\natexlab{}.
\newblock \showarticletitle{Spike-flownet: event-based optical flow estimation with energy-efficient hybrid neural networks}. In \bibinfo{booktitle}{\emph{European Conference on Computer Vision}}. Springer, \bibinfo{pages}{366--382}.
\newblock


\bibitem[Liang et~al\mbox{.}(2024)]%
        {liang2024improving}
\bibfield{author}{\bibinfo{person}{Zhenwen Liang}, \bibinfo{person}{Ye Liu}, \bibinfo{person}{Tong Niu}, \bibinfo{person}{Xiangliang Zhang}, \bibinfo{person}{Yingbo Zhou}, {and} \bibinfo{person}{Semih Yavuz}.} \bibinfo{year}{2024}\natexlab{}.
\newblock \showarticletitle{Improving llm reasoning through scaling inference computation with collaborative verification}.
\newblock \bibinfo{journal}{\emph{arXiv preprint arXiv:2410.05318}} (\bibinfo{year}{2024}).
\newblock


\bibitem[Liu et~al\mbox{.}(2023)]%
        {liu2023improvedllava}
\bibfield{author}{\bibinfo{person}{Haotian Liu}, \bibinfo{person}{Chunyuan Li}, \bibinfo{person}{Yuheng Li}, {and} \bibinfo{person}{Yong~Jae Lee}.} \bibinfo{year}{2023}\natexlab{}.
\newblock \bibinfo{title}{Improved Baselines with Visual Instruction Tuning}.
\newblock


\bibitem[Liu et~al\mbox{.}(2024)]%
        {liu2024improved}
\bibfield{author}{\bibinfo{person}{Haotian Liu}, \bibinfo{person}{Chunyuan Li}, \bibinfo{person}{Yuheng Li}, {and} \bibinfo{person}{Yong~Jae Lee}.} \bibinfo{year}{2024}\natexlab{}.
\newblock \showarticletitle{Improved baselines with visual instruction tuning}. In \bibinfo{booktitle}{\emph{Proceedings of the IEEE/CVF Conference on Computer Vision and Pattern Recognition}}. \bibinfo{pages}{26296--26306}.
\newblock


\bibitem[{OpenAI}(2024)]%
        {openai2024gpt4}
\bibfield{author}{\bibinfo{person}{{OpenAI}}.} \bibinfo{year}{2024}\natexlab{}.
\newblock \showarticletitle{GPT-4 Technical Report}.
\newblock  (\bibinfo{year}{2024}).
\newblock
\urldef\tempurl%
\url{https://arxiv.org/abs/2303.08774}
\showURL{%
\tempurl}


\bibitem[Orchard et~al\mbox{.}(2015)]%
        {orchard2015converting}
\bibfield{author}{\bibinfo{person}{Garrick Orchard}, \bibinfo{person}{Ajinkya Jayawant}, \bibinfo{person}{Gregory~K Cohen}, {and} \bibinfo{person}{Nitish Thakor}.} \bibinfo{year}{2015}\natexlab{}.
\newblock \showarticletitle{Converting static image datasets to spiking neuromorphic datasets using saccades}.
\newblock \bibinfo{journal}{\emph{Frontiers in neuroscience}}  \bibinfo{volume}{9} (\bibinfo{year}{2015}), \bibinfo{pages}{437}.
\newblock


\bibitem[Qu et~al\mbox{.}(2024)]%
        {qu2024e2hqv}
\bibfield{author}{\bibinfo{person}{Qiang Qu}, \bibinfo{person}{Yiran Shen}, \bibinfo{person}{Xiaoming Chen}, \bibinfo{person}{Yuk~Ying Chung}, {and} \bibinfo{person}{Tongliang Liu}.} \bibinfo{year}{2024}\natexlab{}.
\newblock \showarticletitle{E2HQV: High-Quality Video Generation from Event Camera via Theory-Inspired Model-Aided Deep Learning}. In \bibinfo{booktitle}{\emph{Proceedings of the AAAI Conference on Artificial Intelligence}}, Vol.~\bibinfo{volume}{38}. \bibinfo{pages}{4632--4640}.
\newblock


\bibitem[Rebecq et~al\mbox{.}(2019)]%
        {rebecq2019high}
\bibfield{author}{\bibinfo{person}{Henri Rebecq}, \bibinfo{person}{Ren{\'e} Ranftl}, \bibinfo{person}{Vladlen Koltun}, {and} \bibinfo{person}{Davide Scaramuzza}.} \bibinfo{year}{2019}\natexlab{}.
\newblock \showarticletitle{High speed and high dynamic range video with an event camera}.
\newblock \bibinfo{journal}{\emph{IEEE transactions on pattern analysis and machine intelligence}} \bibinfo{volume}{43}, \bibinfo{number}{6} (\bibinfo{year}{2019}), \bibinfo{pages}{1964--1980}.
\newblock


\bibitem[Sobel(1968)]%
        {sobel1968}
\bibfield{author}{\bibinfo{person}{Irwin Sobel}.} \bibinfo{year}{1968}\natexlab{}.
\newblock \bibinfo{booktitle}{\emph{A 3x3 Isotropic Gradient Operator for Image Processing}}.
\newblock \bibinfo{type}{{T}echnical {R}eport} Technical Report. \bibinfo{institution}{Stanford Artificial Intelligence Project (SAIL)}.
\newblock


\bibitem[Son et~al\mbox{.}(2017)]%
        {son20174}
\bibfield{author}{\bibinfo{person}{Bongki Son}, \bibinfo{person}{Yunjae Suh}, \bibinfo{person}{Sungho Kim}, \bibinfo{person}{Heejae Jung}, \bibinfo{person}{Jun-Seok Kim}, \bibinfo{person}{Changwoo Shin}, \bibinfo{person}{Keunju Park}, \bibinfo{person}{Kyoobin Lee}, \bibinfo{person}{Jinman Park}, \bibinfo{person}{Jooyeon Woo}, {et~al\mbox{.}}} \bibinfo{year}{2017}\natexlab{}.
\newblock \showarticletitle{4.1 A 640$\times$ 480 dynamic vision sensor with a 9$\mu$m pixel and 300Meps address-event representation}. In \bibinfo{booktitle}{\emph{2017 IEEE International Solid-State Circuits Conference (ISSCC)}}. IEEE, \bibinfo{pages}{66--67}.
\newblock


\bibitem[Su et~al\mbox{.}(2023)]%
        {su2023event}
\bibfield{author}{\bibinfo{person}{Menghao Su}, \bibinfo{person}{Panpan Yang}, \bibinfo{person}{Runhao Jiang}, {and} \bibinfo{person}{Rui Yan}.} \bibinfo{year}{2023}\natexlab{}.
\newblock \showarticletitle{Event-based object recognition using feature fusion and spiking neural networks}. In \bibinfo{booktitle}{\emph{International Conference on Neural Information Processing}}. Springer, \bibinfo{pages}{470--482}.
\newblock


\bibitem[Tan and Le(2021)]%
        {tan2021efficientnetv2}
\bibfield{author}{\bibinfo{person}{Mingxing Tan} {and} \bibinfo{person}{Quoc Le}.} \bibinfo{year}{2021}\natexlab{}.
\newblock \showarticletitle{Efficientnetv2: Smaller models and faster training}. In \bibinfo{booktitle}{\emph{International conference on machine learning}}. PMLR, \bibinfo{pages}{10096--10106}.
\newblock


\bibitem[Tang et~al\mbox{.}(2024)]%
        {tang2024chain}
\bibfield{author}{\bibinfo{person}{Lv Tang}, \bibinfo{person}{Peng-Tao Jiang}, \bibinfo{person}{Zhi-Hao Shen}, \bibinfo{person}{Hao Zhang}, \bibinfo{person}{Jin-Wei Chen}, {and} \bibinfo{person}{Bo Li}.} \bibinfo{year}{2024}\natexlab{}.
\newblock \showarticletitle{Chain of visual perception: Harnessing multimodal large language models for zero-shot camouflaged object detection}. In \bibinfo{booktitle}{\emph{Proceedings of the 32nd ACM International Conference on Multimedia}}. \bibinfo{pages}{8805--8814}.
\newblock


\bibitem[Tang et~al\mbox{.}(2023)]%
        {tang2023video}
\bibfield{author}{\bibinfo{person}{Yunlong Tang}, \bibinfo{person}{Jing Bi}, \bibinfo{person}{Siting Xu}, \bibinfo{person}{Luchuan Song}, \bibinfo{person}{Susan Liang}, \bibinfo{person}{Teng Wang}, \bibinfo{person}{Daoan Zhang}, \bibinfo{person}{Jie An}, \bibinfo{person}{Jingyang Lin}, \bibinfo{person}{Rongyi Zhu}, {et~al\mbox{.}}} \bibinfo{year}{2023}\natexlab{}.
\newblock \showarticletitle{Video understanding with large language models: A survey}.
\newblock \bibinfo{journal}{\emph{arXiv preprint arXiv:2312.17432}} (\bibinfo{year}{2023}).
\newblock


\bibitem[Wu et~al\mbox{.}(2023)]%
        {wu2023eventclip}
\bibfield{author}{\bibinfo{person}{Ziyi Wu}, \bibinfo{person}{Xudong Liu}, {and} \bibinfo{person}{Igor Gilitschenski}.} \bibinfo{year}{2023}\natexlab{}.
\newblock \showarticletitle{Eventclip: Adapting clip for event-based object recognition}.
\newblock \bibinfo{journal}{\emph{arXiv preprint arXiv:2306.06354}} (\bibinfo{year}{2023}).
\newblock


\bibitem[Yu et~al\mbox{.}(2024)]%
        {yu2024can}
\bibfield{author}{\bibinfo{person}{Zongyou Yu}, \bibinfo{person}{Qiang Qu}, \bibinfo{person}{Xiaoming Chen}, {and} \bibinfo{person}{Chen Wang}.} \bibinfo{year}{2024}\natexlab{}.
\newblock \showarticletitle{Can Large Language Models Grasp Event Signals? Exploring Pure Zero-Shot Event-based Recognition}.
\newblock \bibinfo{journal}{\emph{arXiv preprint arXiv:2409.09628}} (\bibinfo{year}{2024}).
\newblock


\bibitem[Zang et~al\mbox{.}(2024)]%
        {zang2024contextual}
\bibfield{author}{\bibinfo{person}{Yuhang Zang}, \bibinfo{person}{Wei Li}, \bibinfo{person}{Jun Han}, \bibinfo{person}{Kaiyang Zhou}, {and} \bibinfo{person}{Chen~Change Loy}.} \bibinfo{year}{2024}\natexlab{}.
\newblock \showarticletitle{Contextual object detection with multimodal large language models}.
\newblock \bibinfo{journal}{\emph{International Journal of Computer Vision}} (\bibinfo{year}{2024}), \bibinfo{pages}{1--19}.
\newblock


\bibitem[Zheng et~al\mbox{.}(2023)]%
        {zheng2023deep}
\bibfield{author}{\bibinfo{person}{Xu Zheng}, \bibinfo{person}{Yexin Liu}, \bibinfo{person}{Yunfan Lu}, \bibinfo{person}{Tongyan Hua}, \bibinfo{person}{Tianbo Pan}, \bibinfo{person}{Weiming Zhang}, \bibinfo{person}{Dacheng Tao}, {and} \bibinfo{person}{Lin Wang}.} \bibinfo{year}{2023}\natexlab{}.
\newblock \showarticletitle{Deep learning for event-based vision: A comprehensive survey and benchmarks}.
\newblock \bibinfo{journal}{\emph{arXiv preprint arXiv:2302.08890}} (\bibinfo{year}{2023}).
\newblock


\bibitem[Zheng and Wang(2024)]%
        {zheng2024eventdance}
\bibfield{author}{\bibinfo{person}{Xu Zheng} {and} \bibinfo{person}{Lin Wang}.} \bibinfo{year}{2024}\natexlab{}.
\newblock \showarticletitle{EventDance: Unsupervised Source-free Cross-modal Adaptation for Event-based Object Recognition}. In \bibinfo{booktitle}{\emph{Proceedings of the IEEE/CVF Conference on Computer Vision and Pattern Recognition}}. \bibinfo{pages}{17448--17458}.
\newblock


\bibitem[Zhou et~al\mbox{.}(2023)]%
        {zhou2023clip}
\bibfield{author}{\bibinfo{person}{Jiazhou Zhou}, \bibinfo{person}{Xu Zheng}, \bibinfo{person}{Yuanhuiyi Lyu}, {and} \bibinfo{person}{Lin Wang}.} \bibinfo{year}{2023}\natexlab{}.
\newblock \showarticletitle{E-clip: Towards label-efficient event-based open-world understanding by clip}.
\newblock \bibinfo{journal}{\emph{arXiv preprint arXiv:2308.03135}} (\bibinfo{year}{2023}).
\newblock


\bibitem[Zhou et~al\mbox{.}(2024)]%
        {zhou2024eventbind}
\bibfield{author}{\bibinfo{person}{Jiazhou Zhou}, \bibinfo{person}{Xu Zheng}, \bibinfo{person}{Yuanhuiyi Lyu}, {and} \bibinfo{person}{Lin Wang}.} \bibinfo{year}{2024}\natexlab{}.
\newblock \showarticletitle{Eventbind: Learning a unified representation to bind them all for event-based open-world understanding}.
\newblock  (\bibinfo{year}{2024}).
\newblock


\bibitem[Zhu et~al\mbox{.}(2023)]%
        {zhu2023minigpt}
\bibfield{author}{\bibinfo{person}{Deyao Zhu}, \bibinfo{person}{Jun Chen}, \bibinfo{person}{Xiaoqian Shen}, \bibinfo{person}{Xiang Li}, {and} \bibinfo{person}{Mohamed Elhoseiny}.} \bibinfo{year}{2023}\natexlab{}.
\newblock \showarticletitle{Minigpt-4: Enhancing vision-language understanding with advanced large language models}.
\newblock \bibinfo{journal}{\emph{arXiv preprint arXiv:2304.10592}} (\bibinfo{year}{2023}).
\newblock


\bibitem[Zhu et~al\mbox{.}(2024)]%
        {zhu2024multi}
\bibfield{author}{\bibinfo{person}{Jian Zhu}, \bibinfo{person}{Hanli Wang}, {and} \bibinfo{person}{Miaojing Shi}.} \bibinfo{year}{2024}\natexlab{}.
\newblock \showarticletitle{Multi-modal large language model enhanced pseudo 3d perception framework for visual commonsense reasoning}.
\newblock \bibinfo{journal}{\emph{IEEE Transactions on Circuits and Systems for Video Technology}} (\bibinfo{year}{2024}).
\newblock


\end{thebibliography}


\end{document}